# Learning-based Surgical Workflow Detection from Intra-Operative Signals

Ralf Stauder[1,*], Ergün Kayis[1,*], Nassir Navab[1,2]
[1] Computer Aided Medical Procedures, Technical University of Munich, Germany
[2] Department of Computer Science, The Johns Hopkins University, Baltimore, MD, USA
* Joint first authors

**Abstract.** A modern operating room (OR) provides a plethora of advanced medical devices. In order to better facilitate the information offered by them, they need to automatically react to the intra-operative context. To this end, the progress of the surgical workflow must be detected and interpreted, so that the current status can be given in machine-readable form.
In this work, Random Forests (RF) and Hidden Markov Models (HMM) are compared and combined to detect the surgical workflow phase of a laparoscopic cholecystectomy. Various combinations of data were tested, from using only raw sensor data to filtered and augmented datasets. Achieved accuracies ranged from 64% to 72% for the RF approach, and from 80% to 82% for the combination of RF and HMM.

## 1 Introduction

Automatic detection and recognition of the ongoing surgical process is a vital step on the way to a more context-sensitive and collaborative operating room of the future [1]. While it is generally a positive development, that an increasing number of medical and imaging devices are available during surgery, this unfortunately also increases the cognitive workload of the surgeon and the organizational complexity for the OR team. In order to provide the available information only when it is actually necessary for the procedure, devices must recognize the surgical context and workflow.

Different approaches exist in the field of surgical workflow recognition [2], which is also an aspect of the recently defined area of surgical data science. Some methods try to extract a structured model from recorded surgeries [3], while others directly try to recognize the surgical phases or activities through instrument and sensor data [4–6], laparoscopic video [7–9], kinematics information [10], or a mixture thereof [11]. In this work we will apply both Random Forests (RF) [12] and Hidden Markov Models (HMM) [13], separately and combined, to recognize the surgical phase from instrument and sensor data.

## 2    Methods

RF are a collection of randomized decision trees [12]. The individual trees are trained on a randomized subset of all available training data each, and in every node only a randomly selected subset of all available features is available for splitting. This way, all trees of the forest should have a different structure. After training, every tree evaluates each sample and casts a vote, while the majority of votes designates the final classification result. The parameters for the random forest in this experiment were optimized through an exhaustive parameter sweep: The forest consisted of 80 trees with a maximal tree depth of 9 nodes. During training, each node was presented with a random subset with 8 of the 17 available features.

HMMs can be used to estimate and model the internal state of a system based on the trained network structure and recorded observations [13]. A common approach in surgical process modeling is to use left-to-right HMMs to model the surgical workflow, as the internal states usually can be mapped directly to the surgical phases or activities, while the unidirectional structure ensures medically reasonable state transitions. In these experiments the HMM was set up to contain 7 states, corresponding to the surgical phases, which should be detected.

The RF classifier by itself does not take temporal relationships between samples into account, so the classification results can be rather noisy with ambiguous data. Left-to-right HMMs on the other hand are always at risk of switching states too early, or being stuck in wrong phases if no transition event is detected. In an attempt to alleviate these issues, we combined both approaches (see **Fig. 1**). The examined samples are first being classified by the RF. This classification result is then passed to the HMM as observation, and the phase corresponding to the estimated HMM state is given as overall classification result. In this setup, the HMM effectively acts as filter to even out the RF results. The training of this combined setup was done in two steps: First, the RF was trained regularly, using a part of the overall training data. Then the second part of the training data was classified by the trained RF, and the confusion matrix for this classification was generated. This calculated confusion matrix was used directly as emission matrix for the HMM, while the transition matrix was initialized with numbers taken from the ground truth, and later refined through the Baum-Welch-algorithm [14].

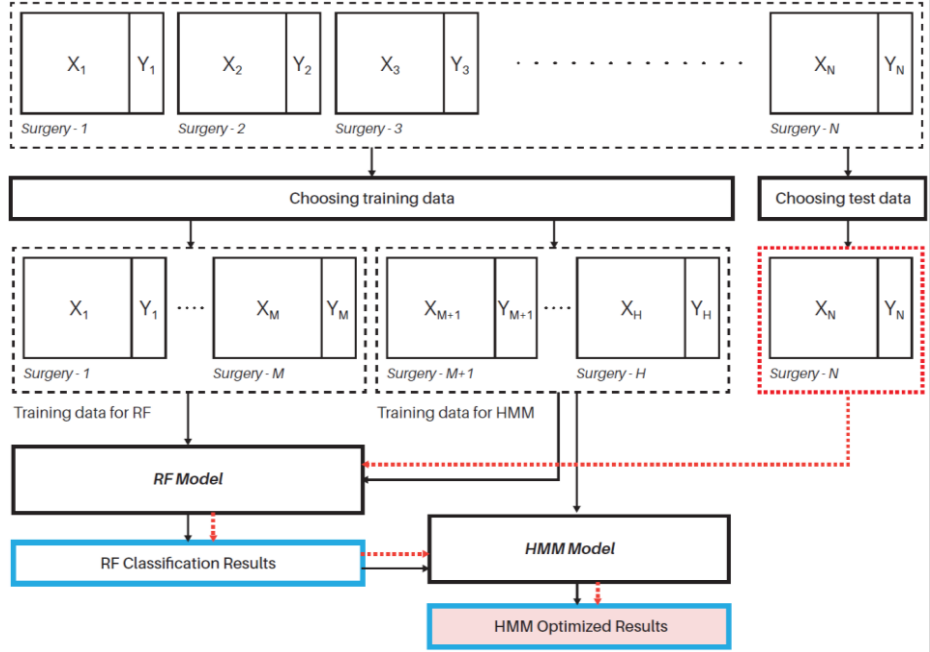

**Fig. 1.** Schematic view of the combined approach between RF and HMM. During training, selected training samples are again split into two parts, so that RF and HMM can be trained separately. For classification, the examined sample is first classified by the RF, then this preliminary result is fed as observation into the HMM.

## 3      Data and Experiments

Two separate, yet related datasets were used in this work. Both were recorded during laparoscopic cholecystectomies, which is the minimally invasive removal of the gallbladder, on real patients in the same hospital, after obtaining appropriate ethics approval. The first dataset, herein called "A", consists of 5 surgeries, of which the laparoscopic video stream, as well as instrument usage and other sensor data were collected. This dataset includes the four recordings used in [6], with one additional surgery added afterwards in the same manner. The second dataset, denoted "B", contains a total of 18 surgeries, although only sensor data, but no video information was stored in these cases. The sensor data includes 12 binary signals, mainly instrument usage [15] and device activation states, 4 analog signals, such as irrigation weight or intra-abdominal pressure, and the time passed (in seconds) since the start of the surgery. The same 7 surgical phase definitions were used in both cases: Trocar placement, Preparation, Clipping, Detaching gallbladder, Retrieving gallbladder, Hemostasis, and Closing. Unless otherwise noted, all of the following experiments were performed on both datasets separately, each time in leave-one-surgery-out fashion, in order to do cross validation.

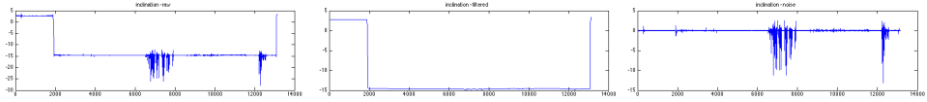

**Fig. 2.** An exemplary plot of a raw, analog signal (left), the filtered, noise-free version of the same signal (middle), and the extracted noise (right).

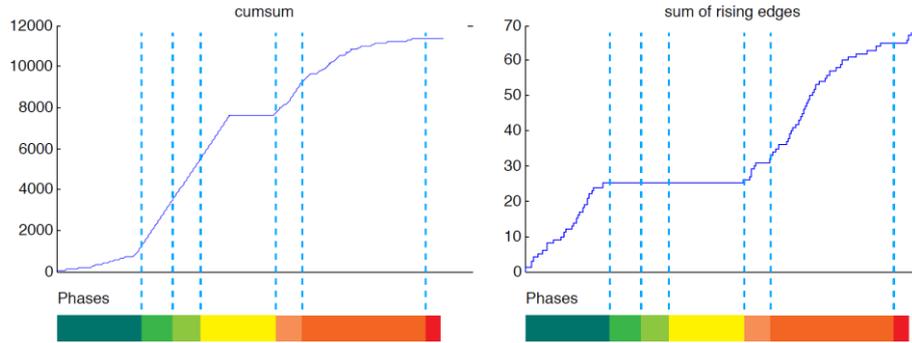

**Fig. 3.** Binary signals were augmented with two auxiliary signals: cumulative sum (left) and the sum of rising edges (right).

For the first experiments, we applied both the RF and HMM to the raw sensor data of each dataset directly. The RF classifier reached an overall accuracy of 69.9% on dataset A, and 70.1% on dataset B, with average Jaccard indices of 60.0% and 57.1% respectively, successfully recreating the setup and results of the work presented in [6]. The HMM achieved an accuracy of 41.8% on the first dataset, and 48.1% on the second, larger dataset, while the average Jaccard indices attained 32.8% and 30.3% each.

As mentioned above, classification of individual frames can be susceptible to signal noise, so in order to improve the signal quality, the data was filtered and augmented for the next experiment. A sliding window median filter over 120 frames was applied to the analog sensor data for smoothing, and the difference of the original and the filtered signals was kept in the dataset as additional noise signal (**Fig. 2**). After processing the first dataset in such a way, the RF achieved an accuracy of 72.0%, and an average Jaccard index of 62.8%. Since median filters do not provide a reasonable filtering on binary signals, these were additionally augmented with two separate, time-dependent features each (**Fig. 3**). The cumulative sum over the binary signal since beginning of the recording is the first additional feature. This feature exhibits a continuous increase in value during usage, flat areas during inactivity, and shallow, unstable increases during noisy periods. The second feature is the sum of rising signal edges since surgery start. This feature has a mainly flat progress during both continuous use and inactivity, but grows rapidly with noisy segments. Training the RF on the datasets with augmentations for analog and digital signals produced mixed results. The overall accuracy dropped for dataset A to 64.0%, but slightly increased to 71.5% on dataset B, while the average Jaccard index improved on both datasets to 65.3% and 60.3% each. The classifications in this case were generally clearer, yet the short phase "Clipping" was completely skipped for some datasets during testing.

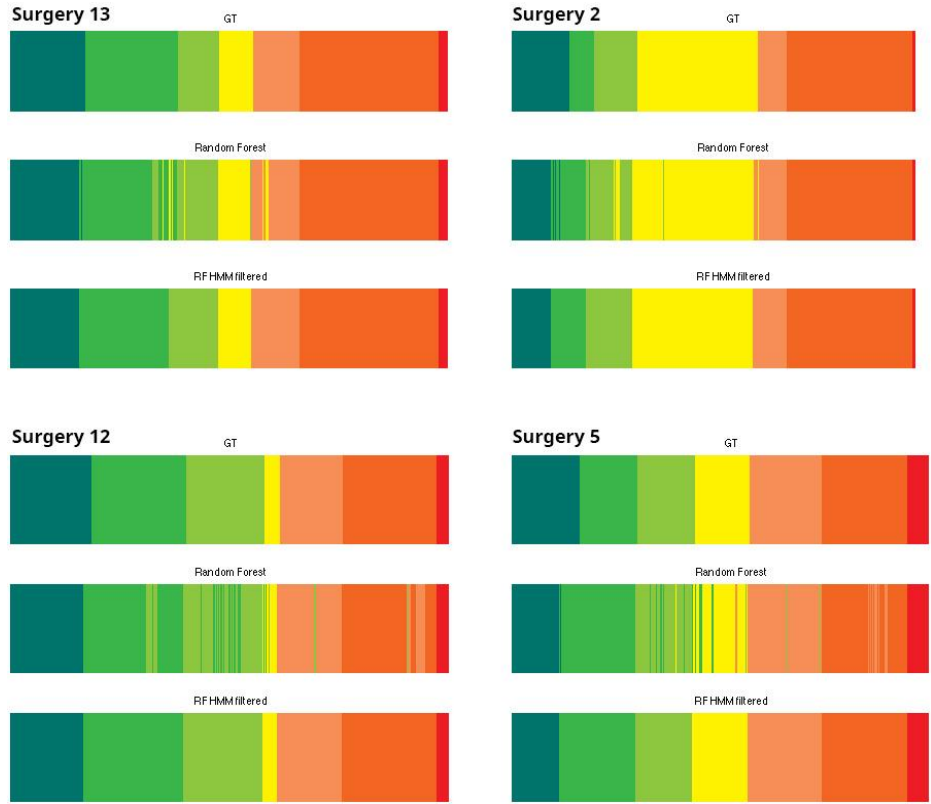

**Fig. 4.** Selected classification results of the combined approach for four surgeries. The top row of each block visualizes the ground truth annotation of the 7 surgical phases over time. The middle row shows the RF classification result, which was then fed as observation to the connected HMM. The bottom row represents the final classification result.

Finally, the two methods of RF and HMM have been combined as described in section 2. Due to the increased requirement for training data, this experiment could not be performed on the first dataset, so all results reported here were achieved on dataset B only. As expected, the output of the combined classifier was more consistent than the pure RF output, yet the phase transitions were more reliable than from the HMM alone (**Fig. 4**). An accuracy of 80.8% was reached on the raw sensor data, and 82.4% after applying the noise-filtering step. The average Jaccard in this case reached 71.1%.

## 4    Discussion

We collected simple, 1-dimensional signals for multiple surgeries, with the goal to detect the surgical workflow only from these signals. We compared the classification quality of Random Forests and Hidden Markov Models, as well as the effect of data augmentation in this limited case. Finally, we connected both classifiers, and were able to notably improve the detection result with this novel, combined approach.

The recognition of surgical phases with HMM based only on such simple signals did not yield reliable results. This can likely be attributed to the fact that the HMM cannot handle the severe signal noise well, and is unable to recover from misclassifications due to the strict left-to-right structure. Also processing the data in order to separate the noise from the expected signal, or providing auxiliary, derived signals only had a minor impact on the classification result. One conceivable explanation for this result is that these derived values were already implicitly represented in the internal decision structure of many random trees. Offering them as separate signal therefore only increased the probability of these values to influence each tree's decision, without actually providing new information.

The combination of RF and HMM succeeded in enhancing the detection result. While the RF has the main influence on the overall classification and phase detection, the HMM positively could filter out erroneous phase changes due to the strict modelling structure.

While video data has also been recorded for some surgeries, it was not yet utilized in this work. Preliminary results suggest that the inclusion of feature data extracted from the video stream can further enhance detection results. Comparing such a system to a more complex, yet possibly slower deep neural network has to be done in future works.